\documentclass[conference]{IEEEtran}
\usepackage{amsmath,graphicx,amssymb,mathtools,dsfont, stfloats}
\usepackage{xcolor}
\usepackage{bm}
\usepackage{algorithm, algpseudocode}
\usepackage{booktabs}
\usepackage{tikz}
\usepackage{pgf}
\usepackage{pgfplots}
\usepackage{balance}
\pgfplotsset{compat=1.18}
\usepackage[caption=false]{subfig}
\usepackage{cite}
\usepackage[hidelinks]{hyperref}

\algnewcommand\algorithmicforeach{\textbf{for each}}
\algdef{S}[FOR]{ForEach}[1]{\algorithmicforeach\ #1\ \algorithmicdo}

\IEEEoverridecommandlockouts

\begin{document}

\newcommand\copyrighttext{%
  \footnotesize \textcopyright 2026 IEEE.  Personal use of this material is permitted.  Permission from IEEE must be obtained for all other uses, in any current or future media, including reprinting/republishing this material for advertising or promotional purposes, creating new collective works, for resale or redistribution to servers or lists, or reuse of any copyrighted component of this work in other works.}
\newcommand\copyrightnotice{%
\begin{tikzpicture}[remember picture,overlay]
\node[anchor=south,yshift=10pt] at (current page.south) {\fbox{\parbox{\dimexpr\textwidth-\fboxsep-\fboxrule\relax}{\copyrighttext}}};
\end{tikzpicture}%
}

\title{ADMM-Based Training for \\ Spiking Neural Networks
\thanks{This work has been supported by the EU through the Horizon Europe/JU SNS project ROBUST-6G (grant no. 101139068) and by the EU under the Italian National Recovery and Resilience Plan (NRRP) Mission~4, Component~2, Investment~1.3, CUP C93C22005250001, partnership on “Telecommunications of the Future” (PE00000001 - program “RESTART”).}
}

\author{\IEEEauthorblockN{Giovanni Perin\IEEEauthorrefmark{1}\IEEEauthorrefmark{2}, Cesare Bidini\IEEEauthorrefmark{1}, Riccardo Mazzieri\IEEEauthorrefmark{1}, and Michele Rossi\IEEEauthorrefmark{1}\IEEEauthorrefmark{3}}
\IEEEauthorblockA{\IEEEauthorrefmark{1} Department of Information Engineering (DEI), University of Padova, Italy}
\IEEEauthorblockA{\IEEEauthorrefmark{2} Department of Information Engineering (DII), University of Brescia, Italy}
\IEEEauthorblockA{\IEEEauthorrefmark{3} Department of Methematics ``Tullio-Levi Civita'', University of Padova, Italy}
giovanni.perin@unibs.it, cesare.bidini@phd.unipd.it, riccardo.mazzieri@phd.unipd.it, michele.rossi@unipd.it
}



\maketitle

\copyrightnotice

\begin{abstract}
In recent years, spiking neural networks (SNNs) have gained momentum due to their high potential in time-series processing combined with minimal energy consumption. However, they still lack a dedicated and efficient training algorithm. The popular backpropagation with surrogate gradients, adapted from stochastic gradient descent (SGD)-derived algorithms, has several drawbacks when used as an optimizer for SNNs. Specifically, the approximation introduced by the use of surrogate gradients leads to numerical imprecision, poor tracking of SNN firing times at training time, and, in turn, poor scalability. In this paper, we propose a novel SNN training method based on the alternating direction method of multipliers (ADMM). Our ADMM-based training aims to solve the problem of the SNN step function's non-differentiability by taking an entirely new approach with respect to gradient backpropagation. For the first time, we formulate the SNN training problem as an ADMM-based iterative optimization, derive closed-form updates, and empirically show the optimizer's convergence, its great potential, and discuss future and promising research directions to improve the method to different layer types and deeper architectures. 
\end{abstract}


%
\section{Introduction}
\label{sec:intro}
Spiking neural networks (SNNs) adopt a neural architecture that closely mimics how the human brain works as an asynchronous, event-based dynamic system. They are especially interesting because they process the input over time, enabling real-time signal processing and inference. Moreover, they show three orders of magnitude of improvement in the energy-delay product (EDP) when running on dedicated hardware, being thus characterized by a great energy efficiency~\cite{rueckauer2022nxtf}. Remarkably, a dedicated training algorithm for SNNs has not yet been developed, and the methods used so far for their supervised training are adaptations of those adopted for traditional NNs~\cite{eshraghian2023training}. A primary drawback with SNNs is the non-differentiable nature of the spiking neuron (Heaviside) activation function, which prevents the direct application of gradient-based optimization methods such as backpropagation. This is often circumvented by using backpropagation with {\it surrogate gradients} \cite{neftci2019surrogate}, where the SNN Heaviside function 
is replaced with a continuous approximation during the backward pass, to enable the computation of its derivative. However, such an approximation prevents an exact tracking of the firing times of SNN neurons at training time, which ultimately impacts the quality of the solution found. This becomes more and more impactful as the number of SNN layers increases. Recently~\cite{fang2021NeurIPS}, researchers have adopted residual connections as a solution to this, but the problem remains; surrogate gradients still provide an approximation to the actual timing behavior of SNN neurons.  

In this work, we propose a fundamentally different and new approach to SNN training. With our method, the activation times of firing neurons are {\it exactly tracked} without the need to use approximations of any sort (e.g., surrogate gradients). In detail, we propose a novel optimization framework specifically tailored for SNNs and based on the alternating direction method of multipliers (ADMM)~\cite{boyd2011distributed}. The learning task is formulated as an optimization problem having the target training loss as its cost function, and defining the SNN neuronal dynamics as its optimization constraints. Our approach is \emph{model-based}, as we optimize a model of the SNN where not only the network weights are learnable (optimized) variables, but also the state variables of the SNN neurons (i.e., the membrane potentials and firing events) are subjected to the learning process. This problem is solved by deriving closed-form updates for the involved variables, with a dedicated subroutine to \emph{optimally handle} the SNN Heaviside function. We stress that this approach completely differs from SGD-derived optimizers, where a forward pass of the data is needed to estimate the value of the loss and the gradient direction before performing an optimization step.

In summary, the contributions of this work are: i) a new optimizer based on the ADMM thought specifically for SNNs is proposed, ii) the optimization problem is relaxed, making it treatable, and closed-form iterative updates with a solid mathematical theory are derived, iii) a subroutine to {\it optimally handle} the SNN Heaviside step function is developed, and iv) a proof-of-concept through numerical simulations to show the potential of the training algorithm is presented. Improvements to the proposed ADMM-based technique are possible and should be pursued to make the method scalable to very large and complex architectures. Such promising directions are discussed in Sect.~\ref{sec:discussion}.
\section{Related Work}
\label{sec:sota}
\subsection{SNNs and training algorithms}
SNNs offer a promising energy-efficient alternative to traditional artificial NNs through sparse, event-driven, and asynchronous computation. 
However, their non-differentiable spike dynamics pose a major challenge for their training. 

The surrogate gradient method~\cite{neftci2019surrogate} is currently the dominant approach for training SNNs. It bypasses the spike function’s discontinuity by introducing a smooth, differentiable surrogate gradient during the backward pass, enabling the application of the backpropagation through time (BPTT) algorithm. Despite its empirical success, this technique comes with several drawbacks: it requires additional state variables and memory, introduces approximation errors that degrade performance, and is also susceptible to vanishing or exploding gradients, as the number of NN layers grows. 

Complementary to gradient-based methods, biologically plausible, gradient-free ``Hebbian'' learning rules have historically received attention in the neuroscience literature. A notable example is spike-timing-dependent plasticity (STDP), a local and biologically inspired learning approach, showing promising results for unsupervised learning and feature extraction~\cite{safa2023fusing}. However, STDP struggles with scalability and accuracy in complex tasks and requires extensive hyperparameter tuning. More recently, forward-only learning algorithms~\cite{nokland2016direct, sigprop2024, hinton2022} aim to bypass the need for backward gradient computation altogether by integrating learning directly during the forward pass.  However, these approaches lack a strong theoretical underpinning, are not yet scalable to deep architectures, and are not competitive with gradient-based alternatives. Additionally, very few explore their use with SNNs and mainly focus on training traditional neural networks. 

Altogether, existing approaches entail trade-offs between biological plausibility, learning effectiveness, and hardware efficiency. This motivates the increasing research interest in alternative training paradigms beyond backpropagation, specifically tailored to the dynamics of SNNs.
\subsection{Gradient-free training with the ADMM}
The use of dual methods and especially the ADMM~\cite{boyd2011distributed} to develop gradient-free optimizers for NNs is rooted in the seminal paper~\cite{taylor2016training}, where the authors show a model-based optimization formulation of a feed-forward NN. The proposed alternating direction solution can be seen as a split Bregman iteration or an inexact formulation of the ADMM. In recent years, the approach gained momentum and has been improved with relaxed approaches to non-convex constraints and a computationally cheap estimation of the pseudoinverse matrix, making the algorithm faster and more efficient~\cite{wang2019admm, wang2020toward, wen2024fast}. Some of these works also prove the convergence of the approach to a stable minimizer. Notably, in~\cite{ebrahimi2025aa} the authors integrate the recent advances with Anderson acceleration, making the ADMM significantly faster and outperforming, among others, advanced SGD-based approaches like Adam and RMSprop. 
In recent years, alternatives to gradient-based methods have also been investigated for recurrent NNs (RNNs)~\cite{tang2021admmirnn, bemporad2023training, adeoye2024inexact}, also using the ADMM. 
For the first time, in this paper, we apply a formulation similar to~\cite{taylor2016training} to the neuronal dynamics of SNNs, which are particular in many aspects (see Sect.~\ref{sec:formulation}). To the best of our knowledge, we are the first to propose a gradient-free optimizer specifically tailored for SNNs and based on the ADMM optimization algorithm.
\section{Optimization problem formulation}
\label{sec:formulation}
Throughout the paper, the layer and time indices will be denoted by $l=1,\ldots,L$ and $t=1,\ldots,T$, respectively. $M$ denotes the number of samples of the dataset (batch), and $n_l$ denotes the number of neurons at layer $l$. The symbol $\Vert \cdot \Vert$ will be used to refer to the Frobenius matrix norm,
and $\langle \cdot \rangle$ will denote the Frobenius inner product.

\begin{figure}
    \centering
    \includegraphics[width=\columnwidth]{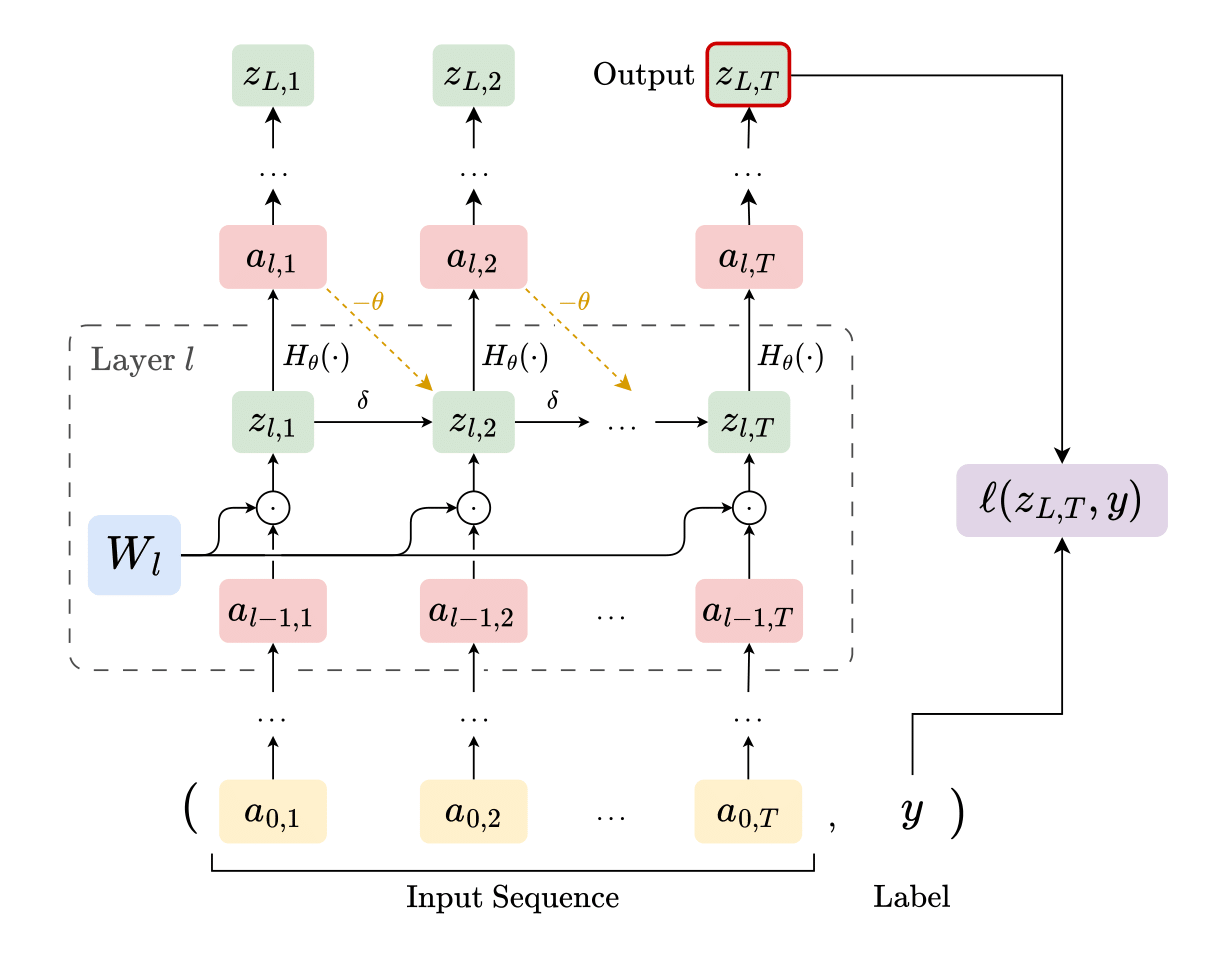}
    \caption{Graphical representation of the SNN model used for the problem formulation in this paper.}
    \label{fig:model}
\end{figure}

The training optimizer is formulated in this section as a model-based optimization problem. Specifically, three sets of matrices (the optimization variables) are defined, namely, i) the model parameters (weights) for each layer $\bm W =\{W_l\}$, where $W_l$ has dimension $n_{l} \times n_{l-1}$ (and the cumulative number of entries of the matrices in $\bm W$ is $\sum_{l=1}^{L} n_l \times n_{l - 1}$, where $n_0$ refers to the number of input neurons); ii) the membrane potentials (pre-activations) for each layer and time $\bm z = \{z_{l,t}\}$, where each $z_{l,t}$ has dimension $n_l \times M$ (and the cumulative number of entries of $\bm z$ is $T\times M\times\sum_{l=1}^{L} n_l$); and iii) the spikes (post-activations) for each layer and time $\bm a =\{a_{l,t}\}$, where again each $a_{l,t}$ has dimension $n_l \times M$ (and the cumulative number of entries of $\bm a$ is $T\times M\times\sum_{l=1}^{L-1} n_l$, as the output layer does not have an activation). The objective function is the target loss. We assume it depends on the membrane potentials of the last layer at the last time index $T$ and on the dataset labels $y$ as $\ell\left(z_{L,T}, y\right)$. The SNN's dynamics are modeled through the constraints relating the optimization variables. This translates to the optimization problem
\begin{equation}
    \begin{aligned}
        \min_{\bm W, \bm z, \bm a} &\quad \ell \left(z_{L,T}, y\right) \\ 
        \text{s.t.} \;\;& \quad z_{l,1} = W_l a_{l-1,1}, &  \forall\, l, \\
        & \quad z_{l,t} = \delta z_{l,t-1} + W_l a_{l-1,t}+\\ &\qquad\quad\;\; - \vartheta a_{l,t-1}, & l \neq L,\; t\neq 1, \\ 
        & \quad z_{L,t} = \delta z_{L,t-1} + W_L a_{L-1,t}, & t \neq 1,  \\
        & \quad a_{l,t} = H_\vartheta (z_{l,t}), & l \neq L,\, \forall\, t, \\
    \end{aligned}
    \label{eq:admm_problem}
\end{equation}
\begin{table*}[bp]
\hrulefill
{\small
\begin{equation}
\begin{aligned}
        \mathcal{L}_{\rho, \sigma}\left(\bm W, \bm z, \bm a, \lambda\right) &= \ell \left(z_{L,T}, y\right) + \frac{\rho}{2}\sum_{l=1}^{L} \left\Vert z_{l,1}-W_la_{l-1,1} \right\Vert^2 + \frac{\rho}{2}\sum_{l=1}^{L-1} \sum_{t=2}^T \left\Vert z_{l,t}-\delta z_{l,t-1}-W_la_{l-1,t}+\vartheta a_{l,t-1} \right\Vert^2\\
        &\quad + \frac{\rho}{2}\sum_{t=2}^T\left\Vert z_{L,t}-\delta z_{L,t-1}-W_La_{L-1,t} \right\Vert^2 + \frac{\sigma}{2}\sum_{l=1}^{L-1} \sum_{t=1}^T \left\Vert a_{l,t} - H_\vartheta(z_{l,t}) \right\Vert^2+\langle z_{L,T}-\delta z_{L,T-1}-W_La_{L-1,T}, \lambda \rangle.
\end{aligned}
\label{eq:relaxed_lagrangian}
\end{equation}
}
\end{table*}
where $a_{0,t}$ are the input training data ($M$ samples). The first constraint models the first time step of every layer, which is an equation similar to the dynamics of a common feed-forward NN: the input is multiplied by the weights matrix to obtain the output. The second constraint encodes the neuronal dynamics for the following time steps at every layer except the last one: the coefficient $\delta$ is the exponential decay factor for the membrane potential, representing the neuron's memory concerning the previous instant, while $\vartheta$ is the firing threshold. If a neuron at layer $l$ fires at time $t$, we have $a_{l,t}=1$ and the membrane potential loses a voltage equal to $\vartheta$ (otherwise ${a_{l,t}}=0$). The third constraint expresses the dynamics of the last layer, where firing and the reset mechanism are disabled. The fourth constraint is the relation between membrane potentials and spikes: a neuron emits a spike when the potential exceeds the threshold $\vartheta$. Note that this can be interpreted computationally as the activation function of the SNN: a Heaviside step function centered in $\vartheta$ (and denoted here by $H_\vartheta$). A pictorial representation of the model is given in Fig.~\ref{fig:model}.
\subsection{Problem relaxation}

Due to the non-convexity of the problem constraints, a relaxed version of problem~\eqref{eq:admm_problem} is derived, where the constraints are added to the objective function as penalties. Following the original approach in~\cite{taylor2016training} and the papers that originated from it, we keep an exact constraint on the dynamics of the last membrane potential. The (relaxed) augmented Lagrangian~\cite{boyd2011distributed} associated with problem~\eqref{eq:admm_problem} is given in Eq.~\eqref{eq:relaxed_lagrangian}, where $\lambda$ is a Lagrange multiplier and $\rho$ and $\sigma$ are parameters tuning the relative weight of the soft constraints concerning the cost function. Intuitively, the Lagrange multiplier for the last membrane potential must be kept because the presence of the objective function adds a drift to the quadratic penalty minimizer. For all the other variables, instead, the sole optimization term is the soft penalty. Therefore, upon convergence, the optimized value of the variables will satisfy the constraints.
\section{ADMM-based solution}
\label{sec:solution}
In this section, we tackle the solution to the relaxed version of the problem via the augmented Lagrangian of Eq.~\eqref{eq:relaxed_lagrangian}, using an alternating direction optimization along the three directions defined by the three sets of optimization variables, plus the dual variable (Lagrange multiplier) update. Without loss of generality, in this work, for the loss function, we use the mean squared error (MSE) between the last membrane potential and the target label, i.e.,
\begin{equation}
    \label{eq:mse}
    \ell(z_{L,T},y) = \Vert z_{L,T}-y\Vert^2.
\end{equation}
This choice is convenient as it leads to a closed-form update. However, closed-form updates can be obtained for many other valid loss functions, and the method can be used even by retrieving the solution to subproblems via numerical solvers when a closed-form solution is not available.

We observe that by minimizing for each variable separately while considering the others as fixed parameters, the Lagrangian in Eq.~\eqref{eq:relaxed_lagrangian} is a convex function. Specifically, each ADMM update solves an unconstrained quadratic program (least squares). This ensures that every update has a simple closed-form solution, which can be easily computed via matrix multiplications (and efficiently implemented on GPUs). Next, we use $\mathds{1}(\cdot)$ as the indicator function, e.g., $\mathds{1}(t<T)$ denotes all time steps smaller than $T$. 
\subsection{Weights update}
We conveniently define the auxiliary tensors 
\begin{align}
    x_l &= [z_{l,1}, z_{l,2} - \delta z_{l,1} + \vartheta a_{l,1}, \ldots,  z_{l,T} - \delta z_{l,T-1} + \vartheta a_{l,T-1}],\nonumber\\
    &\quad\,\, \text{for }l=1, \ldots, L-1, \quad\text{and} \nonumber\\
    x_L &= [z_{L,1}, z_{L,2} - \delta z_{L,1}, \ldots,  z_{L,T} - \delta z_{L,T-1}]. \nonumber
\end{align}
By setting to zero the partial derivatives, the following updates for the weights variables are obtained, as the solution of a linear regression:
{\small
\begin{align}
    W_l &= \left(\sum_{t=1}^T  x_{l,t} a_{l-1,t}^{\scriptscriptstyle{\top}} \right)\left(\sum_{t=1}^T a_{l-1,t} a_{l-1,t}^{\scriptscriptstyle{\top}} \right)^{-1}\,l \neq L,\label{eq:w_update}\\
    W_L &= \left(\frac{1}{\rho} \lambda a_{L-1,T}^{\scriptscriptstyle{\top}}+\sum_{t=1}^T x_{L,t} a_{L-1,t}^{\scriptscriptstyle{\top}} \right)\left(\sum_{t=1}^T a_{L-1,t} a_{L-1,t}^{\scriptscriptstyle{\top}} \right)^{-1}.\label{eq:w_update_L}
\end{align}
}%
\subsection{Pre-activations update}
By minimizing the direction associated with the membrane potential variables $z_{l,t}$ and by momentarily ignoring the presence of the non-differentiable activation function $H_\vartheta$, we find
\begin{align}
    z_{l,t} &= \frac{q_{l,t} + \delta(r_{l,t+1}+\vartheta a_{l,t})\mathds{1}(t<T)}{1+\delta^2\mathds{1}(t<T)} \quad l \neq L, \label{eq:z_update}\\
    z_{L,t} &= \frac{\rho s_{L,t} + \rho \delta r_{L,t+1}\mathds{1}(t<T)}{\rho+\rho\delta^2\mathds{1}(t<T)+2\,\mathds{1}(t=T)}+\nonumber\\
    &\quad+\frac{\left(2y-\lambda\right)\mathds{1}(t=T)+\delta\lambda \mathds{1}(t=T-1)}{\rho+\rho\delta^2\mathds{1}(t<T)+2\,\mathds{1}(t=T)}\label{eq:z_update_L},
\end{align}
where, for convenience, we have defined the auxiliary tensors 
\begin{align}
    p_l &= W_l [a_{l-1,1},\ldots,a_{l-1,T}],\nonumber\\
    s_l &= p_l + \delta [0, z_{l,1}, \ldots, z_{l,T-1}],\nonumber\\
    q_l &= s_l - \vartheta [0, a_{l,1}, \ldots a_{l,T-1}], \quad\text{and}\nonumber\\
    r_l &= - p_l + z_l.\nonumber
\end{align}
Now, we observe that, for $l \neq L$, the terms $\left\Vert a_{l,t} - H_\vartheta(z_{l,t}) \right\Vert^2$ have to be considered (see  Eq.~\eqref{eq:relaxed_lagrangian}).

\begin{table*}[bp]
\hrulefill
{\normalsize
\begin{align}
    a_{l,t} &= \left(\rho W_{l+1}^{\scriptscriptstyle{\top}}W_{l+1}+(\sigma+\rho \vartheta^2 \mathds{1}(t<T))\, I\right)^{-1} \left(\rho W_{l+1}^{\scriptscriptstyle{\top}}v_{l+1,t}-\rho\vartheta w_{l,t+1}\mathds{1}(t<T)+\sigma H_\vartheta(z_{l,t})\right), \quad l \neq L-1, \label{eq:a_update}\\
    a_{L-1,t} &= \left(\rho W_L^{\scriptscriptstyle{\top}}W_L+\left(\sigma +\rho\vartheta^2\mathds{1}(t<T)\right) I\right)^{-1} \left(W_L^{\scriptscriptstyle{\top}} (\rho u_{L,t} + \lambda \mathds{1}(t=T))- \rho\vartheta w_{L-1,t+1}\mathds{1}(t<T)+\sigma H_\vartheta(z_{l,t}))\right) \label{eq:a_update_Lminus1}.
\end{align}
}
\end{table*}
The non-differentiability of $H_\vartheta$ requires a dedicated subroutine that tests the value of the objective function when $H_\vartheta$ is active/inactive (if logic). Specifically, since each entry contributes to the Lagrangian cost in an additive and separable way, we evaluate if commuting the Heaviside step yields a better solution entry-wise (see Algorithm~\ref{alg:z_subroutine}). In Line~\ref{line:commutation_1} the algorithm checks whether $z_{l,t}^{n_l,m} > \vartheta$, causing a spike. However, lowering it below the threshold to prevent firing would yield a better solution, hence we set $z_{l,t}^{n_l,m}$ to the highest possible value not surpassing the threshold (i.e., $\vartheta$ itself), as it is the best solution according to the cost function (a Euclidean norm). The dual case appears in Line~\ref{line:commutation_2}, where the membrane potential $z_{l,t}^{n_l,m}$ is below the threshold $\vartheta$ and generating a spike would yield a better solution. Therefore, the potential is set slightly above the threshold $\vartheta$, using a user-defined and small parameter $\varepsilon > 0$. 
\begin{algorithm}
\caption{z minimizer subroutine}\label{alg:z_subroutine}
\begin{algorithmic}[1]
\Procedure{\texttt{z\_minimizer}}{$z$}
\ForEach{entry of z}
  \If{$z_{l,t}^{n_l,m} > \vartheta$  \& cost$(\vartheta) \le$ cost$(z_{l,t}^{n_l,m})$}
    \State $z_{l,t}^{n_l,m} \gets \vartheta$ 
    \label{line:commutation_1}
  \ElsIf{$z_{l,t}^{n_l,m} \le \vartheta$  \& cost$(\vartheta+\varepsilon) <$ cost$(z_{l,t}^{n_l,m})$}
    \State $z_{l,t}^{n_l,m} \gets \vartheta+\varepsilon$ 
    \label{line:commutation_2}
  \EndIf
\EndFor
\EndProcedure
\end{algorithmic}
\end{algorithm}

\subsection{Post-activations update}
For the post-activations (i.e., the spikes) update, we note that this variable only exists up to and including layer $l=L-1$, as layer $L$ does not have an activation function. For compactness, we define the auxiliary tensors 
\begin{align}
    u_l &= [z_{l,1}, z_{l,2}-\delta z_{l,1}, \ldots, z_{l,T}-\delta z_{l,T-1}],\nonumber\\
    v_l &= u_l + \vartheta [0, a_{l,1}, \ldots, a_{l,T-1}], \quad\text{and}\nonumber\\
    w_l &= u_l - W_l [a_{l-1,1},\ldots,a_{l-1,T}].\nonumber
\end{align}
The spikes $a_{l,t}$ are obtained as given in Eqs.~\eqref{eq:a_update} and~\eqref{eq:a_update_Lminus1}.

Neurons' spikes are physically constrained to be binary variables: either a neuron fires ($a_{l,t} =1$) or it does not ($a_{l,t} =0$). However, projecting the value retrieved in the binary space produces instability for the ADMM iterations towards convergence, as the values might often change abruptly. Therefore, we instead adopt a clipping procedure enforcing $0\le a_{l,t}\le 1$ (i.e., we set $a_{l,t}=\min(\max(0,a_{l,t}), 1)$).

\subsection{Training algorithm}
The full training algorithm using the $z$ minimizer subroutine explained in Algorithm~\ref{alg:z_subroutine} is summarized in Algorithm~\ref{alg:admm}.  
Algorithm~\ref{alg:admm} represents a single ADMM iteration, which must be repeated until a convergence criterion is met~\cite{boyd2011distributed}.
\begin{algorithm}
\caption{ADMM optimizer for SNNs}\label{alg:admm}
\small{
\begin{algorithmic}[1]
\For{$l=1,\ldots,L-1$}
   \State Update the weights $W_l$ with \text{Eq.~\eqref{eq:w_update}} 
   \For{$t=1,\ldots,T$}
    \State Update the pre-activations $z_{l,t}$ with \text{Eq.~\eqref{eq:z_update}}
    \State $z_{l,t} \gets \texttt{z\_minimizer}(z_{l,t})$
    \If{$l < L-1$}
        \State Update the post-activations $a_{l,t}$ with \text{Eq.~\eqref{eq:a_update}}
    \Else
        \State Update the post-activations $a_{L-1,t}$ with \text{Eq.~\eqref{eq:a_update_Lminus1}}
    \EndIf
    \State $a_{l,t} \gets \min(\max(0,a_{l,t}), 1)$
   \EndFor
\EndFor
\State Update the weights $W_L$ with Eq.~\eqref{eq:w_update_L}
\For{$t=1,\ldots,T$}
    \State Update the pre-activations $z_{L,t}$ with \text{Eq.~\eqref{eq:z_update_L}}
\EndFor
\State $\lambda \gets \lambda + \rho(z_{L,T}-\delta z_{L,T-1}-W_La_{L-1,T})$
\end{algorithmic}
}
\end{algorithm}

\section{Numerical simulations}
\label{sec:simulations}

\begin{table}
\centering
\caption{Parameters used for the numerical simulations.}
\label{tab:hyperparams_std}
\begin{tabular}{lc} 
\toprule 
\textbf{Parameter}              & \textbf{Value} \\
\midrule 
Number of time steps $T$ & 150   \\
Neurons per layer $n_l$ & 512 \\
Number of classes $n_c$   & 10    \\
Optimization parameters ($\rho, \sigma, \delta, \vartheta$)         & (1, 0.1, 0.95, 1)     \\
(Total, warming) ADMM iterations & (1000, 300)  \\
Number of training samples $M$    & 200   \\
\bottomrule
\end{tabular}
\end{table}

The simulations were implemented in Python using PyTorch tensors\footnote{Code available at 
\url{https://github.com/cesarbid/SNN_ADMM_Optimizer}} in full compatibility with the existing snnTorch package\footnote{\url{https://snntorch.readthedocs.io/en/latest/}}. In these simulations, whose settings are summarized in Tab.~\ref{tab:hyperparams_std}, we used the neuromorphic dataset N-MNIST. We also adopted a \emph{stochastic ADMM} approach where the order of the updates was chosen randomly for the layers and the time steps, as this procedure yielded better results.

\begin{figure}
    \centering
    \resizebox{\columnwidth}{!}{\input{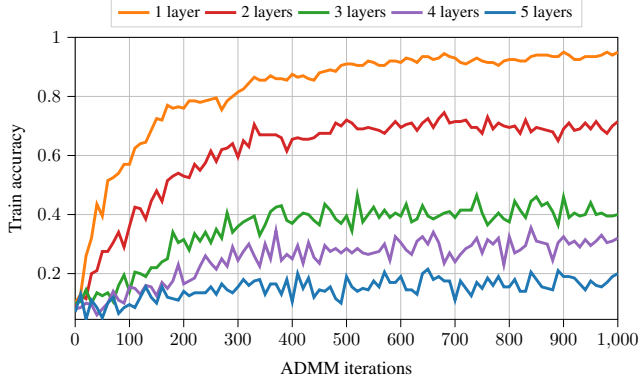}}
    \caption{Train accuracy obtained in 1,000 ADMM iterations varying the number of hidden layers.}
    \label{fig:accuracy_vs_layers}
\end{figure}

\begin{figure}
    \centering
    \subfloat[Lagrangian\label{fig:lagr}]{\resizebox{0.45\columnwidth}{!}{
\begin{tikzpicture}

\definecolor{gainsboro229}{RGB}{229,229,229}
\definecolor{steelblue31119180}{RGB}{31,119,180}

\begin{axis}[
log basis y={10},
tick align=outside,
tick pos=left,
xlabel={ADMM iterations},
xmajorgrids,
xmin=-10, xmax=1000,
xtick style={color=black},
ylabel={$\mathcal{L}_{\rho,\sigma}\left(\bm W, \bm z, \bm a, \lambda\right)$},
ymin=500000, ymax=10000000,
ymode=log,
ymajorgrids,
yminorgrids=true,
grid=both,
ytick style={color=black},
]

\path [draw=steelblue31119180, fill=steelblue31119180, opacity=0.2]
(axis cs:0,7037339.5)
--(axis cs:0,6420657.5)
--(axis cs:10,2063135.625)
--(axis cs:20,1292574.875)
--(axis cs:30,1070081.375)
--(axis cs:40,989167.1875)
--(axis cs:50,951149.8125)
--(axis cs:60,926038.875)
--(axis cs:70,912015.9375)
--(axis cs:80,903878)
--(axis cs:90,895556.1875)
--(axis cs:100,891154.25)
--(axis cs:110,886304.125)
--(axis cs:120,883435.8125)
--(axis cs:130,880321.8125)
--(axis cs:140,878081.625)
--(axis cs:150,876397.75)
--(axis cs:160,874570.125)
--(axis cs:170,873137.8125)
--(axis cs:180,872045)
--(axis cs:190,870811.25)
--(axis cs:200,869965.5)
--(axis cs:210,869042.4375)
--(axis cs:220,868225.25)
--(axis cs:230,867458.625)
--(axis cs:240,866808.625)
--(axis cs:250,866200.9375)
--(axis cs:260,865665.6875)
--(axis cs:270,865198.9375)
--(axis cs:280,864725.0625)
--(axis cs:290,864267.875)
--(axis cs:300,863880.25)
--(axis cs:310,863464)
--(axis cs:320,863110.125)
--(axis cs:330,862782.5625)
--(axis cs:340,862511.5625)
--(axis cs:350,862213.125)
--(axis cs:360,861934.3125)
--(axis cs:370,861654)
--(axis cs:380,861422.3125)
--(axis cs:390,861208.8125)
--(axis cs:400,860985.875)
--(axis cs:410,860774.0625)
--(axis cs:420,860584.6875)
--(axis cs:430,860397.3125)
--(axis cs:440,860201)
--(axis cs:450,860046.3125)
--(axis cs:460,859886.0625)
--(axis cs:470,859714.4375)
--(axis cs:480,859587.5625)
--(axis cs:490,859451.4375)
--(axis cs:500,859295.6875)
--(axis cs:510,859178.6875)
--(axis cs:520,859062.75)
--(axis cs:530,858951)
--(axis cs:540,858822)
--(axis cs:550,858730.375)
--(axis cs:560,858607.875)
--(axis cs:570,858507.6875)
--(axis cs:580,858428.0625)
--(axis cs:590,858330.375)
--(axis cs:600,858231.5)
--(axis cs:610,858139.3125)
--(axis cs:620,858039.375)
--(axis cs:630,857973.5)
--(axis cs:640,857886.125)
--(axis cs:650,857825.0625)
--(axis cs:660,857732.75)
--(axis cs:670,857680.375)
--(axis cs:680,857600.25)
--(axis cs:690,857534.5)
--(axis cs:700,857471.5)
--(axis cs:710,857422.9375)
--(axis cs:720,857349.75)
--(axis cs:730,857281.5625)
--(axis cs:740,857228.5)
--(axis cs:750,857187.6875)
--(axis cs:760,857122.1875)
--(axis cs:770,857083.5625)
--(axis cs:780,857018.125)
--(axis cs:790,856970.1875)
--(axis cs:800,856930.25)
--(axis cs:810,856873.9375)
--(axis cs:820,856826.625)
--(axis cs:830,856784.5)
--(axis cs:840,856745.375)
--(axis cs:850,856708.9375)
--(axis cs:860,856657.4375)
--(axis cs:870,856621.875)
--(axis cs:880,856589.8125)
--(axis cs:890,856541)
--(axis cs:900,856501.5625)
--(axis cs:910,856477.8125)
--(axis cs:920,856446.875)
--(axis cs:930,856409.5)
--(axis cs:940,856374.875)
--(axis cs:950,856338.75)
--(axis cs:960,856297.3125)
--(axis cs:970,856263.75)
--(axis cs:980,856244.6875)
--(axis cs:990,856204.9375)
--(axis cs:1000,856184.0625)
--(axis cs:1000,1032417.5625)
--(axis cs:1000,1032417.5625)
--(axis cs:990,1032476.25)
--(axis cs:980,1032521.125)
--(axis cs:970,1032567.8125)
--(axis cs:960,1032639.5)
--(axis cs:950,1032695.1875)
--(axis cs:940,1032757.4375)
--(axis cs:930,1032815.75)
--(axis cs:920,1032861.5625)
--(axis cs:910,1032931.5)
--(axis cs:900,1033000.25)
--(axis cs:890,1033049.375)
--(axis cs:880,1033117.375)
--(axis cs:870,1033196.8125)
--(axis cs:860,1033259.6875)
--(axis cs:850,1033308.625)
--(axis cs:840,1033388.5625)
--(axis cs:830,1033465.5625)
--(axis cs:820,1033552)
--(axis cs:810,1033614.875)
--(axis cs:800,1033717.0625)
--(axis cs:790,1033799.9375)
--(axis cs:780,1033878.625)
--(axis cs:770,1033937.625)
--(axis cs:760,1034025.8125)
--(axis cs:750,1034126.875)
--(axis cs:740,1034235.5625)
--(axis cs:730,1034336.0625)
--(axis cs:720,1034434.6875)
--(axis cs:710,1034536.1875)
--(axis cs:700,1034618.125)
--(axis cs:690,1034729.5625)
--(axis cs:680,1034859.75)
--(axis cs:670,1034982.9375)
--(axis cs:660,1035095.6875)
--(axis cs:650,1035215.1875)
--(axis cs:640,1035311.3125)
--(axis cs:630,1035469.6875)
--(axis cs:620,1035559.3125)
--(axis cs:610,1035728.875)
--(axis cs:600,1035839.8125)
--(axis cs:590,1036027.4375)
--(axis cs:580,1036173.5)
--(axis cs:570,1036332.75)
--(axis cs:560,1036506.5)
--(axis cs:550,1036631.875)
--(axis cs:540,1036831.9375)
--(axis cs:530,1036969.25)
--(axis cs:520,1037211.4375)
--(axis cs:510,1037403.3125)
--(axis cs:500,1037553.6875)
--(axis cs:490,1037740.8125)
--(axis cs:480,1037961.8125)
--(axis cs:470,1038186.375)
--(axis cs:460,1038438.125)
--(axis cs:450,1038687.375)
--(axis cs:440,1039005.25)
--(axis cs:430,1039235.0625)
--(axis cs:420,1039515.125)
--(axis cs:410,1039885.9375)
--(axis cs:400,1040183.25)
--(axis cs:390,1040516.6875)
--(axis cs:380,1040871.5)
--(axis cs:370,1041237.125)
--(axis cs:360,1041542.375)
--(axis cs:350,1041949.125)
--(axis cs:340,1042356.8125)
--(axis cs:330,1042818.875)
--(axis cs:320,1043376.6875)
--(axis cs:310,1043812.875)
--(axis cs:300,1044354)
--(axis cs:290,1044940.125)
--(axis cs:280,1045689.4375)
--(axis cs:270,1046280.1875)
--(axis cs:260,1047136.9375)
--(axis cs:250,1047936.625)
--(axis cs:240,1048783.125)
--(axis cs:230,1049629.875)
--(axis cs:220,1050812.125)
--(axis cs:210,1051844.375)
--(axis cs:200,1053127.75)
--(axis cs:190,1054581.125)
--(axis cs:180,1056211.25)
--(axis cs:170,1058219.625)
--(axis cs:160,1060347.875)
--(axis cs:150,1062621.875)
--(axis cs:140,1065690.375)
--(axis cs:130,1068838)
--(axis cs:120,1072839.375)
--(axis cs:110,1077744.5)
--(axis cs:100,1083824.625)
--(axis cs:90,1091961.5)
--(axis cs:80,1102230.625)
--(axis cs:70,1116007.25)
--(axis cs:60,1135607.375)
--(axis cs:50,1163759.125)
--(axis cs:40,1214739)
--(axis cs:30,1311808.75)
--(axis cs:20,1543576.5)
--(axis cs:10,2389084.25)
--(axis cs:0,7037339.5)
--cycle;

\addplot [thick, steelblue31119180]
table {%
0 6590027.625
10 2165782.15625
20 1358950.28125
30 1135248.5625
40 1049863.53125
50 1007507.890625
60 982497.984375
70 966845.265625
80 956271.015625
90 948535.96875
100 942381.484375
110 937531.8125
120 933768.78125
130 930541.5625
140 927983.796875
150 925688.125
160 923828.96875
170 922181.40625
180 920751.3125
190 919405.921875
200 918320.84375
210 917273.140625
220 916390.8125
230 915515.84375
240 914763.390625
250 914073.0625
260 913454.296875
270 912840.0625
280 912322.21875
290 911787.578125
300 911331.75
310 910881.09375
320 910494.4375
330 910093.984375
340 909748.96875
350 909415.875
360 909097.1875
370 908815.421875
380 908528.140625
390 908261.546875
400 908001.09375
410 907770.546875
420 907517.625
430 907304.65625
440 907101.09375
450 906892.109375
460 906702.921875
470 906505
480 906346.03125
490 906179.1875
500 906020.25
510 905887.453125
520 905745.25
530 905590.90625
540 905467.234375
550 905334
560 905210.125
570 905094.765625
580 904982.734375
590 904873.640625
600 904747.5625
610 904650.953125
620 904535.859375
630 904458.234375
640 904351.78125
650 904268.453125
660 904180.65625
670 904104.125
680 904008.40625
690 903921.28125
700 903846.765625
710 903777.9375
720 903704.5
730 903626.09375
740 903565.359375
750 903496.578125
760 903421.15625
770 903357.625
780 903306.234375
790 903240.203125
800 903184.3125
810 903116.71875
820 903067.828125
830 903007.359375
840 902957.703125
850 902904.890625
860 902856.21875
870 902809.890625
880 902760.421875
890 902709.609375
900 902662.6875
910 902621.53125
920 902575.890625
930 902535.0625
940 902493.703125
950 902453.34375
960 902403.265625
970 902363.3125
980 902328.203125
990 902291.90625
1000 902255.890625
};
\end{axis}

\end{tikzpicture}}}
    \hspace{2mm}
    \subfloat[Train loss\label{fig:loss}]{\resizebox{0.45\columnwidth}{!}{\input{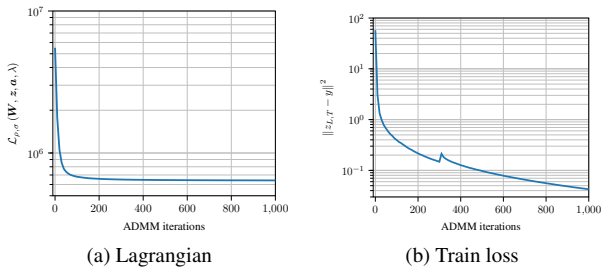}}}
    \caption{Objective functions of the ADMM optimization: relaxed Lagrangian (left) and target loss function (right). Training of the SNN with two hidden layers.}
    \label{fig:obj}
\end{figure}

\begin{figure*}
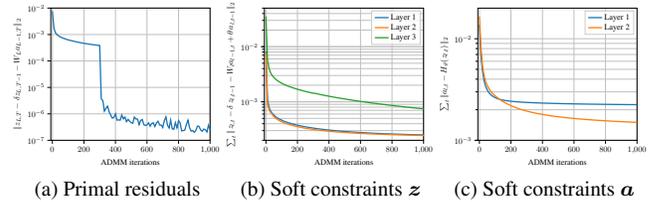

    \centering
    \subfloat[Primal residuals.\label{fig:res}]{\resizebox{0.33\textwidth}{!}{\input{figures/dual_residuals}}}
    \subfloat[Neuronal dynamics soft constraints.\label{fig:sc_z}]{\resizebox{0.33\textwidth}{!}{\input{figures/soft_constraints_z}}}
    \subfloat[Activation soft constraints.\label{fig:sc_a}]{\resizebox{0.33\textwidth}{!}{\input{figures/soft_constraints_a}}}
    \caption{Constraint residuals normalized by their dimensions ($\sqrt{T \times M \times n_{l} }$). Exact constraint primal residual of the last membrane potential (left), norm of the soft constraints relative to the membrane potential dynamics per layer (middle), and norm of the soft constraints relative to the neurons' spikes per layer (right). Training of the SNN with two hidden layers.}
    \label{fig:residuals}
\end{figure*}

In Fig.~\ref{fig:accuracy_vs_layers}, the training accuracy reached for $1,000$ ADMM iterations is shown varying the number of hidden layers and averaging the result across four different runs (shaded areas represent the max-min range of values). As can be seen, the framework is currently solid in solving the classification task with a single hidden layer, with $\sim 98.6$\% accuracy on average, but the performance decreases with an increasing number of hidden layers (around $86$\% accuracy for two hidden layers and progressively lower). The reason is better investigated in the convergence study shown in Figs.~\ref{fig:obj} and~\ref{fig:residuals}, evaluated for the training of the SNN with two hidden layers.

Figure~\ref{fig:lagr} shows the relaxed Lagrangian value during training. From this plot, we see that, after a quick and significant decrease in the first 100 iterations, the Lagrangian reaches a plateau around iteration 200. Hence, it might seem that a stable minimizer has been found; however, by observing the target loss function (Fig.~\ref{fig:loss}), we see that the value is still decreasing and would continue the trend even after iteration $1,000$. Fig.~\ref{fig:residuals} shows the residuals normalized by $\sqrt{T \times M \times n_{l}}$, following a logic similar to the one used to define the stopping criterion for ADMM in~\cite{boyd2011distributed}. Specifically, we observe that the primal residuals (Fig.~\ref{fig:res}) suddenly drop to $10^{-5}$ when the Lagrange multiplier is activated, showing the efficacy of the dual approach. However, the residuals of the soft constraints for the neuronal dynamics (Fig.~\ref{fig:sc_z}) and the activation (Fig.~\ref{fig:sc_a}) decrease at a slower pace: The plateau reached by the soft constraints of the hidden layers hence dominates the contribution to the relaxed Lagrangian of Fig.~\ref{fig:lagr}. This causes a \emph{mismatch between the optimization and the prediction at inference time}, since the neuronal dynamics are not respected with enough numerical precision, and activation variables $\bm a$ are not exactly in $\{0,1\}$. The effect becomes more evident as the number of layers and time steps increases. Allowing the soft constraints to further reduce their values with more training iterations would improve the performance.

\section{Discussion}
\label{sec:discussion}
The current approach shows potential, but there is still room for improvement. We note that, unlike classic NNs, SNNs have a time dimension that adds a factor of the order of $10^3$ in the number of variables, making the problem more complex to tackle. Nonetheless, the ADMM is known for handling problems with up to $\sim 10^9$ variables when several empirical techniques are combined with the standard framework~\cite{boyd2011distributed}, as the convergence might be extremely sensitive to hyperparameters, as we also observed in our experiments. For instance, as also done in other works, an adaptive value for $\rho$ and $\sigma$ across iterations and the addition of Anderson acceleration~\cite{ebrahimi2025aa} are expected to speed up the convergence of the soft constraints. These enhancements would also make the optimizer more scalable concerning the number of layers and time steps. 

In the present work, we demonstrated that this first approach effectively handles the presence of the non-differentiability of the Heaviside step function, as the dedicated subroutine developed produces a stable (although slow) convergence towards the minimizer, representing a significant novelty concerning the commonly used surrogate gradient methods. We also note that the approach can be extended to convolutional SNN layers by modifying the soft constraints relative to the membrane potential dynamics. In this case, the multiplication between the weight matrix $W_l$ and the spikes $a_{l-1,t}$ would be replaced by a correlation operation, which is still linear and thus can be handled by the ADMM. It is also important to observe that, like in previous approaches derived from~\cite{taylor2016training}, the proposed training algorithm is suitable for processing the entire dataset, instead of processing small batches in series and performing many (imprecise) backpropagation steps with a coarse estimation of the gradient, as SGD-based optimizers do. In sharp contrast, with the ADMM, fewer optimal subproblems are executed at each iteration, spanning, in principle, the whole dataset. We additionally note that the available memory can be a bottleneck in this case. The solution to this is the consideration that variables $z_{l,t}$ and $a_{l,t}$ are independent sample-wise: hence, several CPUs and/or GPUs \emph{can process different subsets of the dataset in parallel}. However, the weights update requires integrating the information with the update
\begin{equation}
    \label{eq:parallel_w_update}
    W_l = \left(\sum_{n=1}^N\sum_{t=1}^T x_{l,t}^n a_{l-1,t}^n{^{\scriptscriptstyle{\top}}}\right)\left(\sum_{n=1}^N\sum_{t=1}^Ta_{l-1,t}^n a_{l-1,t}^n{^{\scriptscriptstyle{\top}}}\right)^{-1},
\end{equation}
where $n$ is the worker (similarly, batch) index. Remarkably, this directly enables federated learning (FL), where the parameter server (PS) computes the reduction update in Eq.~\eqref{eq:parallel_w_update}, while the clients keep a portion of the dataset and update the variables $z_{l,t}^n$ and $a_{l,t}^n$ locally and privately.
\section{Concluding remarks}
\label{sec:conclusions}
In this work, we presented a first formulation of an ADMM optimizer specifically tailored for SNN training. The iterative solution only uses closed-form updates and a subroutine with if-else logic to optimally handle the presence of the non-differentiable Heaviside step function. Notably, the proposed solution solves the poor approximation issue of backpropagation with surrogate gradients. With this work, we aim to start a new research line on SNN optimizers: We believe that the proposed method has great potential, and there exists significant room for improvement in terms of the type of layers supported, convergence speed, memory utilization, and scalability to large networks.

\bibliographystyle{IEEEbib}
\bibliography{refs}

\end{document}